\documentclass[review]{elsarticle}

\usepackage{amssymb}
\usepackage{latexsym}
\usepackage{hyperref}
\usepackage{graphicx}
\usepackage{subfigure}
\journal{Journal of \LaTeX\ Templates}

\begin{document}

\begin{frontmatter}

\title{Class agnostic moving target detection by color and location prediction of moving area\tnoteref{tnote1}}%

\author[label1]{Zhuang He}

\author[label1]{Qi Li\corref{cor1}}
\cortext[cor1]{Corresponding author:}
\emailauthor{liqi@zju.edu.cn}{Qi Li}

\author[label1]{Huajun Feng}

\author[label1]{Zhihai Xu}

\address[label1]{The State Key Laboratory of Modern optical Instruments, Zhejiang University, No.38 Zheda Road, Hangzhou, China}

\begin{abstract}
Moving target detection plays an important role in computer vision. However, traditional algorithms such as frame difference and optical flow usually suffer from low accuracy or heavy computation. Recent algorithms such as deep learning-based convolutional neural networks have achieved high accuracy and real-time performance, but they usually need to know the classes of targets in advance, which limits the practical applications. Therefore, we proposed a class agnostic moving target detection algorithm. This algorithm extracts the moving area through the difference of image features. Then, the color and location probability map of the moving area will be calculated through Bayesian estimation. And the target probability map can be obtained through the dot multiply between the two maps. Finally, the optimal moving target area can be solved by stochastic gradient descent on the target probability map. Results show that the proposed algorithm achieves the highest accuracy compared with state-of-the-art algorithms, without needing to know the classes of targets. Furthermore, as the existing datasets are not suitable for moving target detection, we proposed a method for producing evaluation dataset. Besides, we also proved the proposed algorithm can be used to assist target tracking.
\end{abstract}


\end{frontmatter}

\section{Introduction}
\label{sec1}
Moving target detection plays an important role in computer vision, it has wide applicated value in domains as security monitoring, human-computer interaction and automatic pilot. Recently, the integration of multiple computer vision tasks is often required in intelligent systems\citep{multitask}. And the moving object detection is usually used as the front end of many other computer vision tasks such as target tracking, behavior analysis and object re-identification. Therefore, we argue that the rapid implementation of moving target detection is very important. To detect a moving target in two frames or a small number of frames, can make the system have sufficient time to analyze, identify, track or record the target.

Traditional moving target detection algorithms include background elimination\citep{backgroundMOG1}, optical flow\citep{opticalflow} and frame difference \citep{framedifference}. The background elimination algorithm needs to model the background of the scene. The optical flow algorithm needs to match all of the pixels between the images. These two algorithms are time-consuming and not suitable for the requirements of real-time moving target detection. For the frame difference algorithm, it is fast but easily affected by camera shaking, background clutter and noise. Besides, this method is difficult to detect the entire target completely. Recently, convolution neural networks based on deep learning have made great progress in target detection. Typical algorithms are video salient detection and video object detection. Video salient detection is to identify the most visually unique targets or areas in the videos and then segment them from the background\citep{vsd3, vsd1}. Such methods detect targets based on the semantic information but ignore the motion information. And they sometimes extract static targets with visually salient, which do not match the expectations of the moving target detection. Video object detection is to identify the class and location of the targets in the videos\citep{tcnn,vod1} . Although many of object detection methods have made great progress in speed and accuracy. They need to know the possible classes of the detected targets in advance, which means they are not class agnostic. Therefore, they are still limited in practice application. Especially in scenes where the targets are unknown, such as security monitoring, target tracking and segmentation.

Many computer vision tasks domains as tracking and  security monitoring are more interested in moving targets than statics targets. While the state-of-the-art moving target detection algorithms are less to extract the motion features, which is the most unique characteristic of moving targets.\citep{zhang2019portraitnet, le2018video, cheng2016video,li2015extracting} In this work, we propose a class agnostic moving target detection algorithm. Firstly, the proposed algorithm extracts the moving area based on the difference of convolution neural network features. Secondly, it computes the color features and location features on the moving area. Through estimating the Bayesian estimation of the colors belong to the moving target, the moving target probability map can be obtained. Finally, the proposed algorithm generates an initial detection box based on the moving area, and optimizes it by stochastic gradient descent (SGD) to obtain the optimal moving target area. The proposed algorithm makes full use of the motion characteristics of moving targets, and can detect moving targets effectively.

For the evaluation process. As mentioned above, we hope that the algorithms can achieve the detection of moving targets in two frames. Therefore, we need the datasets to provide pairs of adjacent images for moving target detection. And the targets should be annotated in the form of box, which is convenient for the propagation between different computer vision tasks. However, we find there is less datasets specially for moving target detection algorithm evaluation. The datasets for video salient detection\citep{fbms, davis,segtrackv2} are annotated at pixel-level, which do not match the required annotation form. The datasets for video object detection\citep{ilsvrc2015} are annotated at box-level. However, the classes of the targets are relatively fixed, and it is easy to bias the evaluation results. The tracking datasets\citep{otb2013, otb2015,vot2017,vot2018} meet the needs of moving object detection to some extent. They are all from the nature videos whose target classes are relatively free, and the targets are annotated at box-level. The tracking datasets consist of consecutive video frames, which is convenient for us to extract image pairs for moving target detection. However, there are still some videos in the datasets do not match the requirements. For example, some videos are annotated by part of the targets rather than the whole targets, and some videos contain unannotated moving targets. Such videos should be removed when selecting evaluation dataset. Therefore, we propose a method to produce dataset for moving target detection based on tracking dataset. The produced dataset contains more than 1500 image pairs annotated at box-level. Which can make a good evaluation for moving target detection algorithms. 

We compare the proposed algorithm with some state-of-the-art algorithms on the produced dataset. Results show that the proposed algorithm can detect the motion area accurately and extract the moving target by given its bounding box. It runs fast and achieves high accuracy without needing to know the classes of targets in advance. Additionally, we also apply the proposed algorithm as a module to assist target tracking. Results show the accuracy of the assisted trackers are all improved. We argue that the proposed algorithm will be widely used in practice.

To summarize, the main contributions are fourfold:
\begin{itemize}
	\item We proposed a fast and class agnostic algorithm for moving target detection.
	\item We proposed a method for generating the moving target detection dataset.
	\item We evaluated the proposed algorithm and other state-of-the-art algorithms on the produced dataset. Results show our algorithm achieved the highest detection accuracy.
	\item The proposed algorithm can be used as a module to assist target tracking to improve the tracking accuracy.
\end{itemize}

\section{Related works}
Moving target detection is a hot topic in computer vision for many years. And many algorithms and researches have been proposed. In this section, we give a brief overview of recent works on target detection in two lines: traditional moving target detection algorithms, and convolutional neural networks based on deep learning. In addition, we also give a brief overview of the datasets for target detection.

\subsection{Traditional moving target detection algorithms}
Typical traditional moving target detection algorithms are frame difference algorithms, background elimination algorithms and optical flow algorithms. The frame difference algorithms\citep{framedifference} extract the changed pixels through the difference between image pairs. Then, the moving area can be obtained by binarizing the difference image. The original frame difference method is easily affected by camera shaking, background clutter, illumination variation and noise. Many methods and tricks have been proposed to improve the robustness of frame difference algorithms. For example, adopt adaptive binary threshold to reduce the impact of illumination change, and introduce morphological filtering to remove noise. Although many works have been done, the algorithms based on frame difference are still unstable and low accuracy for moving target detection.
Background elimination algorithms have been widely used in security monitoring. Typical algorithms are MOG\citep{backgroundMOG1, backgroundMOG2} and GMG \citep{backgroundGMG}. Such algorithms generate the background model based on the first few frames. In the detection process, the moving targets can be obtained by subtracting the model from the input image frames. Background elimination algorithms are also easily disturbed by noise, background change and illumination change. In addition, the process of generating background model is usually time-consuming, and some algorithms with model update \citep{backgroundGMG} are even difficult to achieve real-time in detection process. 
Optical flow was first proposed by Horn and Schunck\citep{opticalflow}. It represents the projection of the actual motion of a three-dimensional object on the image plane. Which can be simply understood as the velocity vector field of the target. Due to its high accuracy, it has been widely used in motion detection. While the optical flow needs to match all pixels of two images, and is very time-consuming in practice, especially for the images with millions of pixels. Although some algorithms \citep{pointopticalplow1} have proposed to extract few of feature points for calculating optical flow, the complexity of the algorithm is still heavy. 

\subsection{Moving target detection algorithms based on convolutional neural networks}
With the development of deep learning, algorithms based on convolutional neural networks (CNN) have made great progress in moving target detection. Typical algorithms are video object detection and video salient detection. Video object detection is to detect the classes and locations of targets in the videos. They have developed rapidly since 2015, benefitting from the development of image object detection algorithms. The state-of-the-art image object detection networks such as Alex-net\citep{alexnet}, RCNN\citep{rcnnn}, Fast RCNN\citep{fastrcnn}, Faster RCNN\citep{fasterrcnn}, YOLO\citep{yolo} and SSD\citep{ssd} have been well applied in video object detection. Typical algorithms are T-CNN \citep{tcnn}, \citep{vodRCNN} and \citep{RCNN}. The general strategy of such algorithms is to detect the targets in the first few frames based on the image object detection, and determine the position of the targets in the subsequent frames through tracking-by-detection, such as correlation filter, siamese net and flow-net\citep{vodflow}. The video object detection algorithms can identify the class and position of the target in the entire video sequences accurately, and run in real-time. However, they only introduce the motion features of the moving targets in the tracking process, but completely depends on the image object detection in the detection process. This makes the video object detection algorithms can only detect targets whose classes are known in advance. Therefore, they are less suitable for moving target detection in practice, especially for the unknown targets.

The researches on video salient detection have also made great progress these years. Early salient detection algorithms extract semantically significant targets in the images. Such as \citep{staticsaliency1}, \citep{staticsaliency2}, and \citep{staticsaliency3}. Which do not focus on motion features of the moving targets the same as video object detection algorithms. With the increasing popularity of video processing in computer vision, the salient detection is also applied in videos. Wang proposed a fully convolutional networks for video salient object detection\citep{vsod}. This algorithm extracts the static saliency of single image firstly. Then, the image pairs and the static saliency image are fed into the video salient object detection network to extract the video salient areas. This algorithm can effectively extract the salient targets in videos, and runs with the speed of 2fps on GPU. While its detection results extremely rely on the static salient detection, and perform less accuracy on motion estimation. Recently, the flow-net have been widely used in video salient detection and made great progress \citep{vsdflows}. The flow-net can provide a wealth of motion information of the images.  The image pairs are fed into the networks and the moving target probability map is output directly. Such algorithms achieved high accuracy and stability, which are suitable for moving target detection. However, these algorithms are relatively complex and usually have high requirements on equipment. Which is very limited in practical applications.
\subsection{Datasets for moving target detection}
We find the existing datasets are not suitable for the evaluation of moving target detection algorithms. FBMS \citep{fbms} and DAVIS \citep{davis} are often used in video salient detection, which are originally produced for motion segmentation. They contain 59 and 50 natural video sequences, respectively. And are fully-annotated at pixel-level for each frame. Although this annotation is very accurate, it is not conducive to the description and recording of the targets. We argue that the box-level annotation is more suitable for moving target detection. It can simplify the records of the target, and is easier for the propagation between different tasks, such as tracking and person re-identification. ILSVRC\citep{ilsvrc2015} is one of the most popular dataset in computer vision. It can be used in many computer vision tasks such as classification, object detection and segmentation. The ILSVRC contains millions of images and are annotated in multiple levels. However, the classes of the targets in the dataset is fixed and easy to learn. This will bias the evaluation results of the class agnostic moving target detection algorithm. OTB \citep{otb2013, otb2015} and VOT \citep{vot2017,vot2018}are the most commonly used tracking datasets. They consist of hundreds of videos which contain the moving targets in various challenge situations. Fully-annotated box-level ground truth for each frame is available. And the classes of targets are more than the classes of targets in ILSVRC. However, it is not reasonable to use the tracking datasets directly for moving target detection. One reason is some sequences are only annotated by part of the targets rather than the whole targets, and some videos contain unannotated moving targets. The other reason is the similarity of consecutive video frames is too high, which is easy to bias the evaluation accuracy.

In this work, we illustrated the importance of motion features in moving target detection. Then, we propose a fast and class agnostic moving target detection algorithm based on color and location prediction of motion area. The proposed algorithm can detect moving targets effectively. It is computationally efficient, much faster than many traditional models and other deep learning networks. Besides, we also proposed a method for producing dataset for moving target detection algorithms. We evaluate the proposed algorithm on the produced dataset. Results show the proposed algorithm achieves the highest accuracy compared with the previous algorithms. Furthermore, we applied the proposed algorithm as a module to assist target tracking. The results show the accuracy and robustness of the assisted trackers are all improved.

\begin{figure*}[htbp]
	\centering
	\includegraphics[scale=0.3]{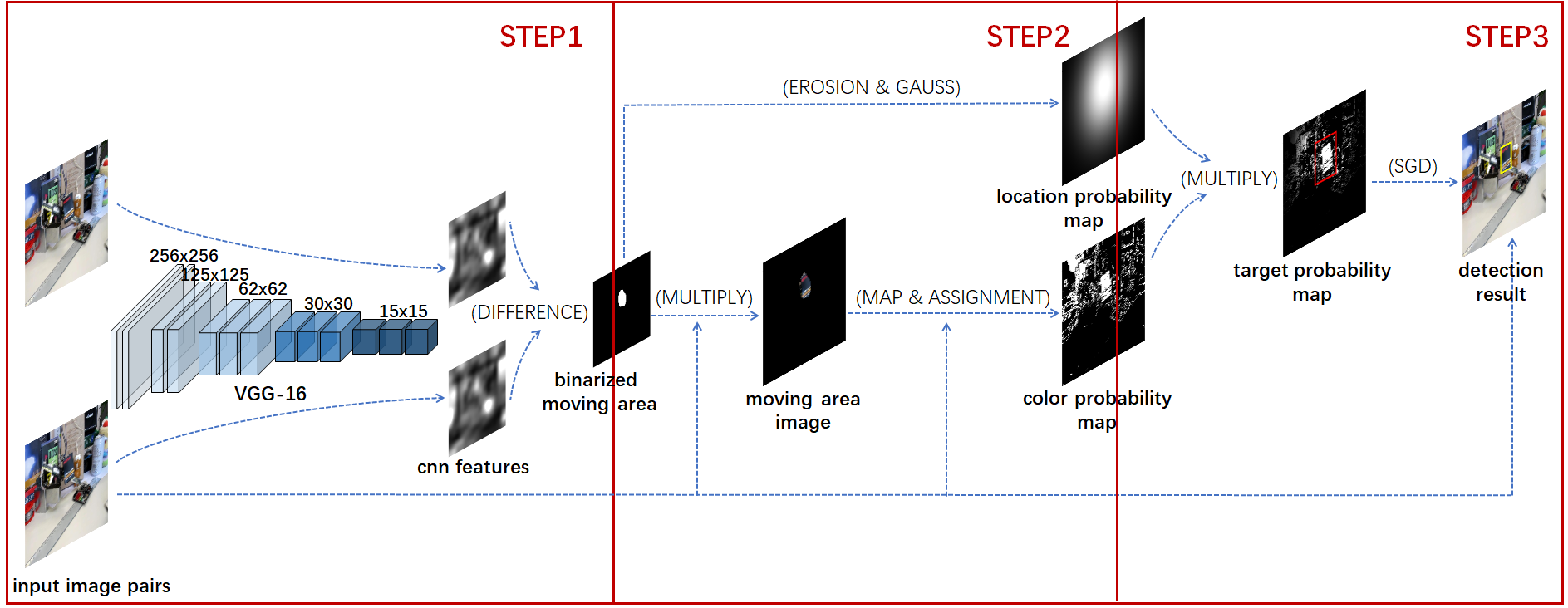}
	\caption{The pipeline of the proposed algorithm. Where the uppercase letters represent the operation process, and the lowercase letters represent the images obtained in the process.}
	\label{Fig.1}
\end{figure*}

\section{Proposed algorithm}

The pipeline of the proposed class agnostic moving target detection algorithm as shown in Fig.\ref {Fig.1} . It can be divided into three steps. The STEP1 calculates the binarized moving area through the difference between the CNN features of image pairs. In STEP2, the color and location prediction of the moving area is solved. Then, the color probability map and location probability map can be obtained through Bayesian estimation estimation and Gaussian distribution generation, respectively. In STEP3, the target probability map is calculated through the dot multiply between the color probability map and location probability map. And the optimal moving target bounding box can be solved through stochastic gradient descent (SGD) based on the target probability map. 

\subsection{Extract moving area}
Motion feature is the inherent feature of the moving targets. Extract stable and accuracy motion feature is important for moving target detection. The difference between consecutive video frames can reflect the motion characteristics of pixels. However, the traditional frame difference algorithms which based on the raw data of image pairs is unstable and easily affected by noise. We found that the deep CNN features of the image are not sensitive to noise, illumination change, and small vibrations. Therefore, we proposed a method for extracting moving area based on image CNN feature difference, as shown in Fig.\ref {Fig.1} STEP1. The process can be expressed as
\begin{equation}
I_{dff}=\varphi\left(I_{1}\right)-\varphi\left(I_{2}\right)
\end{equation}
Where $I_{1}$ and $I_{2}$ are the inputs of the algorithm, which are adjacent image pairs. The inputs should be three-channel images. If the inputs are gray images, they should be copied into three-channel images. $\varphi(\cdot)$ represents the process of extracting CNN features. The CNN features in this algorithm is Vgg16-layer14. $I_{dff}$  represents the differenced image, and it will be accumulated in the channel dimension to get two-dimensional difference image $I_{dff2}$. Then, the binarized moving area can be extracted through
\begin{equation}
I_{tho}=f\left(I_{dff2}, tho\right)
\end{equation}
Where $f(\cdot)$  represents the binarization of the difference image, and $tho$  is the binarization threshold, which is 0.8 times of the maximum pixel value of $I_{dff2}$. $I_{tho}$  is the binarized moving area, as shown in Fig.\ref {Fig.1} STEP1. Although the resolution of the binarized moving area image is 16 times lower than the original image, we are just to extract the approximate moving area without using it for precise positioning. The obtained binarized moving area can still be used to estimate the motion features effectively.

\subsection{Extract moving features}
\label{sec3.2}
The color and location characteristics of the binarized moving area can be used to distinguish the moving target from the background. In STEP2, the binarized moving area image is upsampled to the original image size through cubic interpolation algorithm. The moving area image can be obtained through the dot multiply between the upsampled binary image and input image $I_{2}$, as shown in Fig.\ref {Fig.1} STEP2. Then, the color histogram statistics is made on the moving area image and the input image, respectively.

\begin{equation}
H_{mot}=hist\left(I_{mot}\right)
\end{equation}

\begin{equation}
H_{2}=hist\left(I_{2}\right)
\end{equation}
Where $hist(\cdot)$  represents the process of color histogram statistics, each channel of the color is equally divided into 16 parts, and the length of the color histogram is 16 *16 * 16 = 4096. $I_{mot}$ is the moving area image.  $H_{mot}$ is the color histogram of the moving area image and $H_{2}$  is the color histogram of $I_{2}$ . Through Bayesian estimation, the probability that each color belongs to the target can be solved. The formula is 

\begin{equation}
\mathop{H_{target}(c)}\limits_{c=1:4096} = P(c | mot) \times P(mot) / P(c)
\end{equation}
Where $c$  is the index of the color histogram.  $H_{target}(c)$ represents the probability that  $c$  belongs to the moving target.  $P(c | mot)$ represents the probability of  $c$ appearing in moving area image $I_{mot}$ . Which can be solved through

\begin{equation}
P(c | mot)=\frac{H_{mot}(c)}{\sum\limits_{l=1:4096} H_{mot}(l)}
\end{equation}
$P(mot)$ represents the probability of moving area appearing in the input image, which can be solved through the ratio between the number of non-zero pixels in $I_{mot}$  and $I_{2}$ .   $P(c)$ represents the probability of $c$  appearing in the input image $I_{2}$ . Which can be solved through
\begin{equation}
P(c)=\frac{H_{2}(c)}{\sum\limits_{l=1:4096} H_{2}(l)}
\end{equation}
The $H_{target}$  is a 4096-length vector whose value represents the probability that the corresponding color belongs to the moving target. Then, the input image $I_{2}$ is processed pixel-by-pixel. Its color histogram index $c$ is calculated and reassigned by the corresponding probability $H_{target}(c)$ . Finally, the color probability map can be obtained, as shown in Fig.\ref {Fig.1} STEP2.

In addition to the color characteristics, we argue that the moving area can also reflect good location confidence. Therefore, we calculate the center of the  moving area image as follows: Erode the moving area through a circular template with diameter 1, and repeat this step until only 1 pixel remains. The coordinate of the center pixel is represented as  $(x,y)$ . Then, the location probability map can be obtained through generating a Gaussian distribution centered on the center pixel, as shown in Fig.\ref {Fig.1} STEP2. The standard deviation of the Gaussian distribution is 1/5 of the image size in this work.

\subsection{Detect moving target}
\label{sec3.3}
The obtained moving probability map and color probability map can reflect the color and location characteristics of moving targets. Through the dot multiply between the two maps, the target probability map can be obtained, as shown in Fig.\ref {Fig.1} STEP3. Its pixel value indicates the probability that it belongs to the moving target. We argue that the box containing more high-value pixels and fewer low-value pixels is more likely to be the moving target area. Based on the binarized moving area, an initial box is generated on the target probability map. As shown in the red box in the target probability map in Fig.\ref {Fig.1} STEP3. The center of the initial box is the same as the center of the Gaussian distribution in \ref {sec3.2} , which is $(x,y)$ . And the width and height of the box can be calculated by the following equation.
\begin{equation}w=h=\sqrt{2 N}\end{equation}
Where $w$  and  $h$ represent the width and height of the box respectively, $N$  is the number of the non-zero pixels of the upsampled binarized motion area. Then, the size and location of the initial box will be optimized through stochastic gradient descent (SGD). Details as follows.

\begin{equation}
x=l r \cdot \frac{S(x, y, w, h)-S(x+\Delta x, y, w, h)}{\Delta x}
\end{equation}
\begin{equation}
y=l r \cdot \frac{S(x, y, w, h)-S(x, y+\Delta y, w, h)}{\Delta y}
\end{equation}
\begin{equation}
w=l r \cdot \frac{S(x, y, w, h)-S(x, y, w+\Delta w, h)}{\Delta w}
\end{equation}
\begin{equation}
h=l r \cdot \frac{S(x, y, w, h)-S(x, y, w, h+\Delta h)}{\Delta h}
\end{equation}
Where $lr$  is the learning rate.  $\Delta x$, $\Delta y$, $\Delta w$ and $\Delta h$ represent the disturbance of the box.  $S(\cdot)$ is the scoring function of the box. It can be calculated by
\begin{equation}
S(x, y, w, h)=\frac{{sum}\left(I_{target},(x, y, w, h)\right)}{w \cdot h}+\lambda \cdot \frac{w+h}{W+H}
\end{equation}
Where  $I_{target}$  represents the target probability map. ${sum}\left(I_{target},(x, y, w, h)\right)$ represents the sum of the pixels value in the box $(x, y, w, h)$  in $I_{target}$  .  $W$ and  $H$ represent the width and height of the input image, respectively. The first term is to maximize the average probability of the box. The second term is a penalty term to prevent the box from being too small and detecting part of the target.  $\lambda$ is the coefficient of the penalty term. This process can be accelerated with integral image. Repeat the above iteration process until the box $(x, y, w, h)$  is unchanged, or the number of iterations reaches 100. The result box of the moving target detection algorithm can be obtained, as shown in the yellow box in Fig.\ref {Fig.1} STEP3.

Compared with the previous algorithms, the advantage of the proposed algorithm is that it makes good use of the target's motion characteristics. It can be well used to detect moving targets, especially for targets with unknown classes.

\section{Evaluation}

\subsection{Datasets}
The exiting datasets for video salient detection and video object detection are not suitable for moving target detection. Therefore, we proposed a method for producing dataset for the evaluation of moving target detection algorithms based on tracking datasets. Typical tracking datasets consist of multiple videos with hundreds of frames, such as OTB100 \citep{otb2015} and VOT2017 \citep{vot2017}. They contain many classes of targets, and the scenes are from the reality. However, the tracking datasets still have some limitations in the evaluation of moving target detection algorithms. The highly similar scenes in the video sequences reduce the diversity of the datasets. Which will bring a lot of redundant calculations and bias the evaluation results. Besides, some videos such as the ‘Biker’ and ‘Boy’ in OTB100 are annotated by part of the moving target. Furthermore, some videos contain more than one moving target and the annotated targets are not the most significant moving targets, such as ‘Bolt’ and ‘Crowds’. In addition, in order to make the dataset more similar to reality, we extract image pairs at random intervals to simulate different camera frequencies. Therefore, we proposed the following three rules for producing the datasets.

\begin{itemize}
	\item Image pairs are selected every 50 frames to reduce the scene similarity, and the interval between the image pairs ranges from 1 to 10 randomly. For example, the frame numbers of the selected image pairs are: \{[1,3], [51, 55], [101, 109], [151, 153], …\}.
	\item Manually reannotate targets which are not fully annotated.
	\item Remove videos with multiple moving targets.
\end{itemize}

\begin{figure*}[htbp]
	\centering
	\subfigure[]{
		\centering
		\includegraphics[scale=0.4]{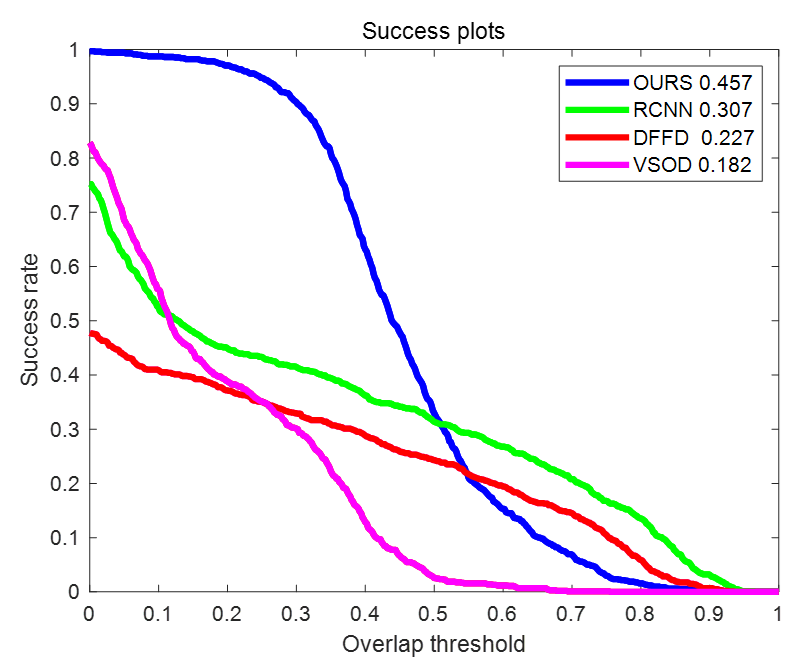}
		\label{Fig.2a}
	}%
	\subfigure[]{
		\centering
		\includegraphics[scale=0.4]{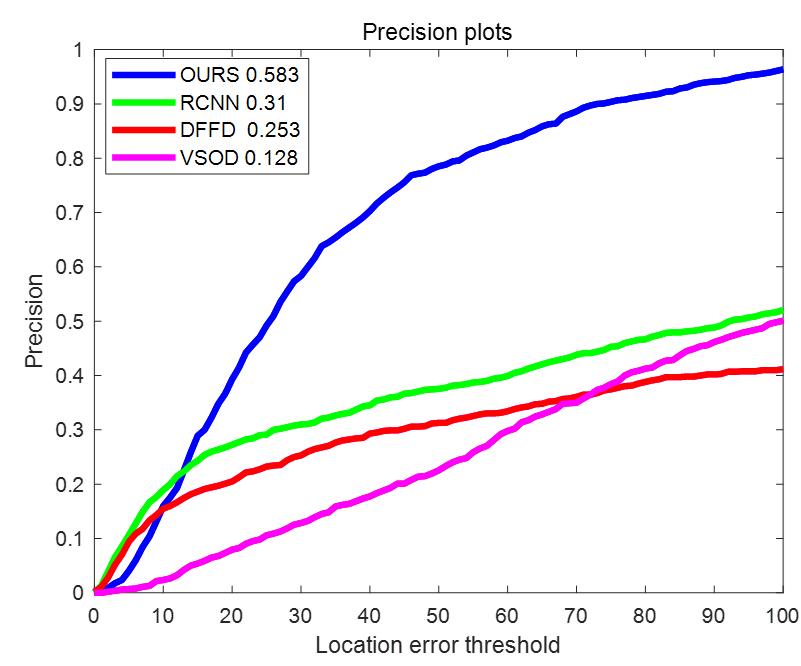}
		\label{Fig.2b}
	}%
	\centering
	\caption{The evaluation results of the four algorithms on the produced dataset. Where OURS represents the proposed class agnostic moving target detection algorithm. RCNN, DFFD and VSOD represent the algorithm proposed by \citep{RCNN}, \citep{DFFD} and \citep{vsod}, respectively.}
	\label{Fig.2}
\end{figure*}

We extract image pairs on 110 different sequences on OTB100 and VOT2017 based on the three rules. The produced dataset contains more than 1500 image pairs. It is fully annotated at box-level, and covers a wide camera frequencies. The images in the produced dataset are completely from the natural world. The evaluation on it can reflect the algorithms’ performance in practical applications effectively.

\subsection{Evaluation of the proposed algorithm}

In this experiment, we compare the detection accuracy of the proposed algorithm with some state-of-the-art algorithms, which are video object detection based on Faster rcnn \citep{RCNN}, video object detection based on flow-net \citep{DFFD}, and video salient detection by fully convolution neural network \citep{vsod}. For the two video object detection algorithms, as they may detect multiple boxes, we select the box which most overlaps with the binarized moving area as the result. Where the binarized moving area is extracted the same as Fig.\ref {Fig.1} STEP1. For the video salient detection algorithm, as the result is pixel-level, we get the optimal bounding box of the salient area through SGD the same as Fig.\ref {Fig.1} STEP3. 
The intersection over union (IOU) between the results and ground truth is used to evaluate the detection accuracy. It can be calculated by
\begin{equation}I O U=\frac{R \cap G}{R \cup G}\end{equation}
Where $R$  and $G$  represent the result box and ground truth, respectively.  ${R \cap G}$ represents the intersection area between  $R$  and $G$ , and ${R \cup G}$  represents the union area between  $R$  and $G$  . IOU can reflect the overlap between the results and ground truth effectively. Its value varies between 0 and 1. The result box with large IOU indicates that it is close to the ground truth. Besides, the distance between the center of result and ground truth is also used to estimate the position accuracy of the algorithms. The proposed algorithm and three state-of-the-art algorithms are evaluated on the produced dataset, the results as shown in Fig.\ref {Fig.2} . Where OURS represents the proposed class agnostic moving target detection algorithm. RCNN, DFFD and VSOD represent the algorithm proposed by \citep{RCNN}, \citep{DFFD} and \citep{vsod}, respectively. Fig.\ref {Fig.2a} shows the success rate of the algorithms under different IOU threshold. And the upper right corner shows the area under curve (AUC) of the algorithms which is the area enclosed by the curves and the x-y axis. Fig.\ref {Fig.2b} shows the precision of the algorithms under different location error threshold, and the precision at the threshold of 30 pixels is shown at the upper left corner. 

\begin{figure}[htbp]
	\centering
	\includegraphics[scale=0.6]{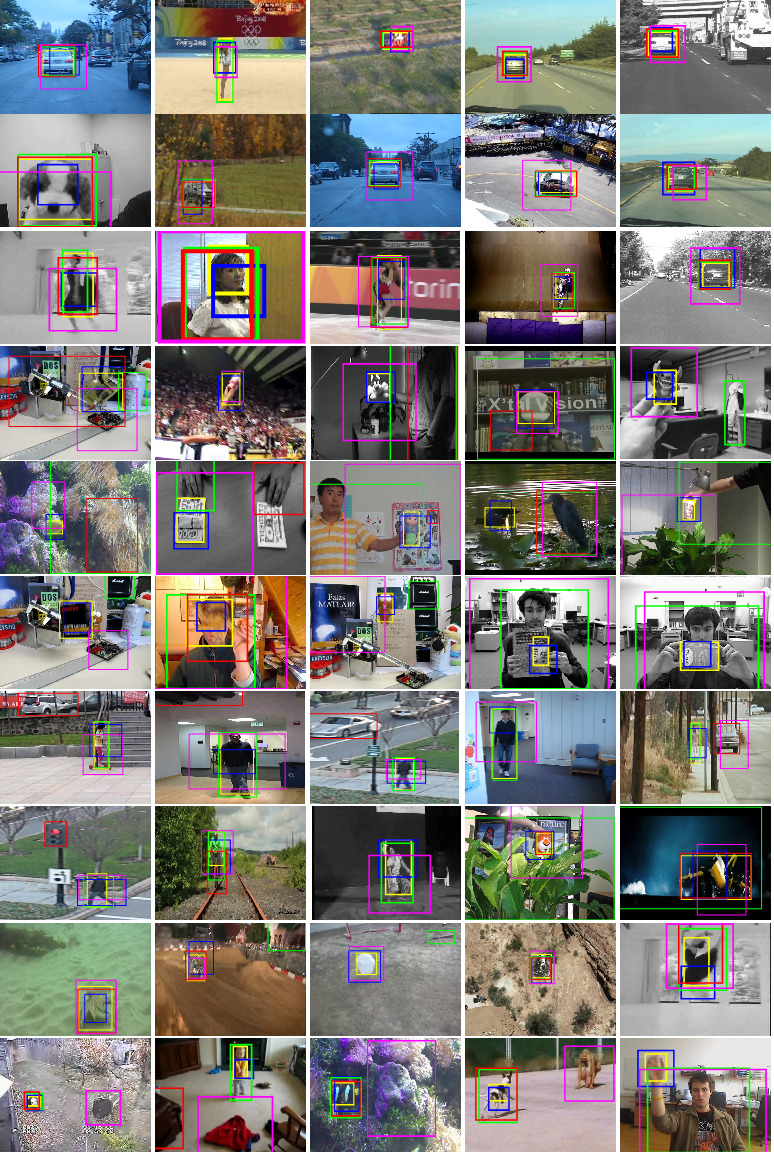}
	\caption{Comparison of the detection results of the proposed algorithm, RCNN, DFFD, VSOD and the ground truth. Their detection results are represented by blue, green, red, purple and yellow boxes, respectively.}
	\label{Fig.3}
\end{figure}

It can be found that the proposed algorithm achieves the highest accuracy on the produced dataset. Its AUC of the success rate is 0.457, which is higher than the AUC of RCNN, DFFD and VSOD by 0.15, 0.23 and 0.275, respectively. Its precision at the threshold of 30 pixels is 0.583, which is higher than the precision of RCNN, DFFD and VSOD by 0.273, 0.33 and 0.455, respectively. Comparing with RCNN and DFFD, we can find that the accuracy of the proposed algorithm is higher under low overlap threshold, while lower under high overlap threshold. The reason is that the detection results of RCNN and DFFD extremely depend on the image object detection. When the classes of the moving targets are contained in their training dataset, high accuracy of the detection results can be obtained. Otherwise, the accuracy of the detection results is low, and even wrong targets will be detected. The difference between the success rate of RCNN and DFFD is also because the target classes contained in their training datasets is different. The classes in RCNN are 'background', ’aeroplane', 'bicycle', 'bird', 'boat', 'bottle', 'bus', 'car', 'cat', 'chair', 'cow', 'diningtable', 'dog', 'horse', 'motorbike', 'person', 'pottedplant', 'sheep', 'sofa', 'train' and 'tvmonitor'. While the classes in DFFD are 'airplane', 'antelope', 'bear', 'bicycle', 'bird', 'bus', 'car', 'cattle', 'dog', 'domestic-cat', 'elephant', 'fox', 'giant-panda', 'hamster', 'horse', 'lion', 'lizard', 'monkey', 'motorcycle', 'rabbit', 'red-panda', 'sheep', 'snake', 'squirrel', 'tiger', 'train', 'turtle', 'watercraft', 'whale' and 'zebra'. The classes in RCNN appear more in the produced dataset, which makes its AUC is 0.08 higher than DFFD. For the comparation between the proposed algorithm and VSOD, it can be found that the success rate of the proposed algorithm is better than VSOD under all overlap rate thresholds. We argue the reason is that the detection results of VSOD depend on the static salient detection. The produced dataset is all from the nature videos, which are generally complicated, such as background clutter, occlusion, and non-moving target interference. While the static salient detection algorithms are usually difficult to get good results in the above scenes. Therefore, the detection accuracy of VSOD is usually low.  Fig.\ref {Fig.3} shows more intuitive results of these algorithms on the produced dataset. It can be found that these algorithms can detect moving targets effectively in most cases. For the targets whose classes have been pre-trained in RCNN and DFFD, the accuracy of RCNN and DFFD even higher than the proposed algorithm, as shown in Fig.\ref {Fig.3} lines 2-3. While for the targets whose classes have not been pre-trained, the RCNN and DFFD usually lose the moving targets, as shown in Fig.\ref {Fig.3} line 4-6. As the pre-trained classes of RCNN and DFFD are not the same, their performance on the dataset is different. For example, when the target class is ‘person’, RCNN performs good, while DFFD often loses the target. When the target class is ‘tiger’, DFFD performs good, while RCNN loses the target. The results as shown in Fig.\ref {Fig.3} lines 7-8. For the moving targets whose background is simple, VSOD can obtain good detection results, as shown in Fig.\ref {Fig.3} line 9. However, VSOD tends to lose the moving target when the background is clutter, or the moving target is not salient, as shown in Fig.\ref {Fig.3} line 10.

The evaluation results show that video object detection algorithms and video salient detection algorithms both have limitations in the class agnostic moving target detection process. The reliance on object detection and saliency detection makes them difficult to detect untrained target classes in complex scenes. Different from the above algorithms, the algorithm proposed in this paper starts with the inherent motion characteristics of the targets, and extracts the color and location features of the moving target, makes the detection of moving targets more accurately. The proposed algorithm is not trained on the evaluation dataset and runs in real-time with 40fps on i7-9700CPU and 1060Ti GPU. Therefore, the proposed algorithm can be well applied in reality.

\subsection{Ablation analysis}

In this part, we analyze the effect of each module in the proposed algorithm. Four kinds of features are evaluated on the frame difference process. And the effect of the modules of binarization (BIN), color probability confidence (CPC), location probability confidence (LPC) and SGD is also evaluated. Results as shown in Table.\ref {tab1}.  Where Layer3 and Layer14 represent the CNN feature of Vgg16-layer3 and Vgg16-layer14, respectively. HOG represents the histogram of oriented gradients, which is proposed by \citep{hog}. RAW represents the raw data of the image. Success Rate (SR) represents the AUC of the IOU curve, and PRE (30) represents the precision at the location error threshold at 30 pixels.

\begin{table*}[!t]
	\caption{\label{tab1}Evaluation of the effect of each module on moving target detection}
	\centering
	\begin{tabular}{p{2cm}lllllllllll}
		\hline
		Methods  & 0   & 1   & 2   & 3   & 4   & 5   & 6   & 7   & 8   & 9   & 10  \\
		\hline
		Layer3   &    &    &    & \checkmark  &    & \checkmark  & \checkmark   &       &     &     &     \\
		Layer14  &       &       & \checkmark     &       &       & \checkmark     & \checkmark     & \checkmark     & \checkmark     & \checkmark     & \checkmark  \\
		HOG      &       &       &       &       & \checkmark    &       & \checkmark    &       &       &       &       \\
		RAW      & \checkmark     & \checkmark     &       &       &       &       &       &       &       &       &       \\
		BIN      & \checkmark     & \checkmark     & \checkmark    & \checkmark     & \checkmark    & \checkmark     & \checkmark    & \checkmark    & \checkmark     & \checkmark     & \checkmark     \\
		LPC      &       & \checkmark     & \checkmark     & \checkmark     & \checkmark     & \checkmark    & \checkmark    &       &       & \checkmark     & \checkmark     \\
		CPC      &       & \checkmark     & \checkmark     & \checkmark     & \checkmark     & \checkmark     & \checkmark     &       & \checkmark     &       & \checkmark     \\
		SGD      &       & \checkmark     & \checkmark     & \checkmark     & \checkmark     & \checkmark     & \checkmark    & \checkmark     & \checkmark     & \checkmark     &       \\
		\hline
		SR       & 0.042 & 0.066 & 0.457 & 0.155 & 0.093 & 0.167 & 0.169 & 0.351 & 0.442 & 0.430 & 0.172 \\
		\hline
		PRE (30) & 0.098 & 0.122 & 0.583 & 0.296 & 0.173 & 0.313 & 0.310 & 0.422 & 0.548 & 0.553 & 0.178\\
		\hline
	\end{tabular}
\end{table*}

Method 0 differences the image pairs based on the raw data of the images, and extracts the moving target area through binarization. It can be viewed as the traditional frame difference method. Its SR is 0.042 and PRE (30) is 0.098. Which shows that the traditional frame difference method is difficult to achieve high accuracy for the moving target detection in realistic complex scenes. Methods 1, 2, 3 and 4 show the impact of different features on the detection of moving target. Results show that methods 2, 3, and 4 with robust features perform better than method 1 with raw data of the image. And the method 2 with Vgg16-layer14 performs the highest accuracy, whose SR is 0.457 and PRE (30) is 0.583. It shows that the deep features are more advantage for moving target detection. In addition, we also performed evaluations on multiple features combination, and the results are shown in methods 5 and 6. The comparison of methods 2, 5 and 6 shows that multiple features even perform worse than single deep feature. The reason is the shallow features with high-resolution are easily affected by background and camera shaking. Therefore, in the multiple features fusion stage, shallow features often introduce a lot of noise, which will make the detection results worse. Method 7 uses SGD to solve the moving target on the binarized moving area image directly, without extracting the characteristics of the moving area. Its SR and PRE (30) are decreased by 23.2\% and 27.6\% compared to method 2. It shows that the motion features are important for moving targets detection. Method 8 does not estimate the LPC map, and performs the SGD on the color probability map directly. Its SR and PRE (30) are 0.442 and 0.548, respectively. Compared to method 7, the result shows CPC can bring 25.9\% improvement to SR and 29.9\% improvement to PRE (30), respectively. Method 9 does not estimate the CPC map, it uses the dot multiply of the differenced image and LPC map as the target probability map. Its SR and PRE (30) are 0.430 and 0.553. Compared to method 7, the result shows LPC can bring 22.5\% improvement to SR and 31.0\% improvement to PRE (30), respectively. In method 10, the SGD optimization is not adopted, the initial box solved by \ref {sec3.3}  is used as the detection result directly. Its SR and PRE (30) are decreased by 62.4\% and 69.5\% compared to method 2. which shows that the SGD optimization can greatly improve the accuracy of moving targets detection. 

The results of ablation analysis show that all the modules of the proposed algorithm play an important role in moving target detection. We also find that the features have the greatest impact on the detection of moving targets. The reason is that features can directly affect the extraction of binarized moving areas. Thereby affecting the color probability map and location probability map. In addition to the features, the optimization of the bounding box is also very important. Good optimization module can improve the overlap between the result box and target very much. Therefore, we argue that the features and optimization methods is particularly important for moving target detection. The features which do not sensitive to background motion and camera shaking, and optimization methods with good convergence can improve the detection accuracy greatly.

\subsection{Evaluation on the target tracking}
We argue that the proposed algorithm can also be used in other computer vision tasks, such as target tracking\citep{tracking}. The general form of target tracking is to find the position of the target in the video frames based on the given target in the first frame. The target is usually represented by a rectangular box in tracking. In this work, we make three changes to the proposed algorithm and apply it as a module to assist target tracking.

\begin{itemize}
	\item Detect the moving target in the searching area the same as the trackers instead of the entire image. This change can make the proposed algorithm integrate with the tracking algorithms better and improve the calculation efficiency.
	\item Calculate the color probability map based on the box obtained in the previous frames instead of the binarized moving area. This change can make the estimation of the color probability map more accurate.
	\item Use the tracking results in the current frame as the initial box, and optimize it on the target probability map. As many tracking algorithms are provided with powerful scale estimation modules, we only modify the position of the box in SGD.
\end{itemize}

\begin{figure}[!t]
	\centering
	\subfigure[]{
		\centering
		\includegraphics[scale=0.4]{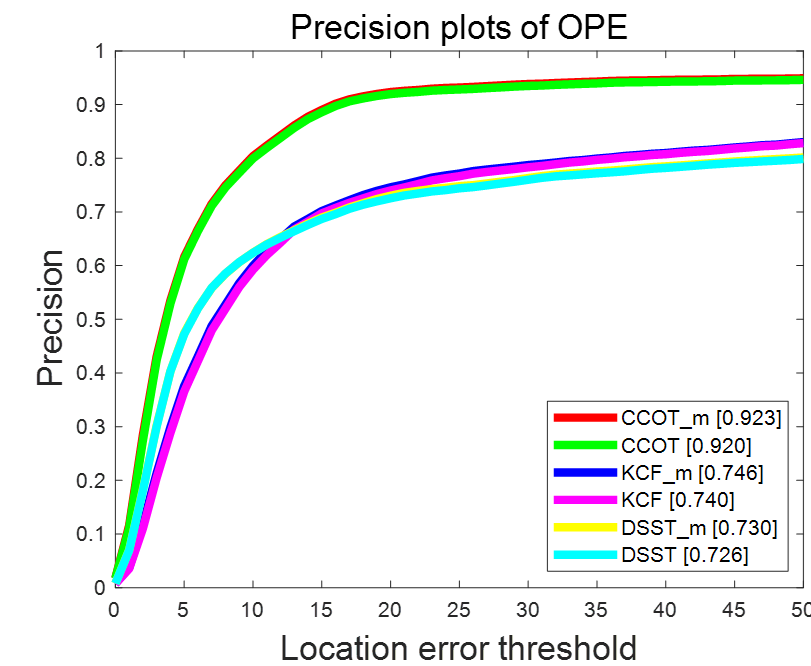}
		\label{Fig.4a}
	}%
	\subfigure[]{
		\centering
		\includegraphics[scale=0.4]{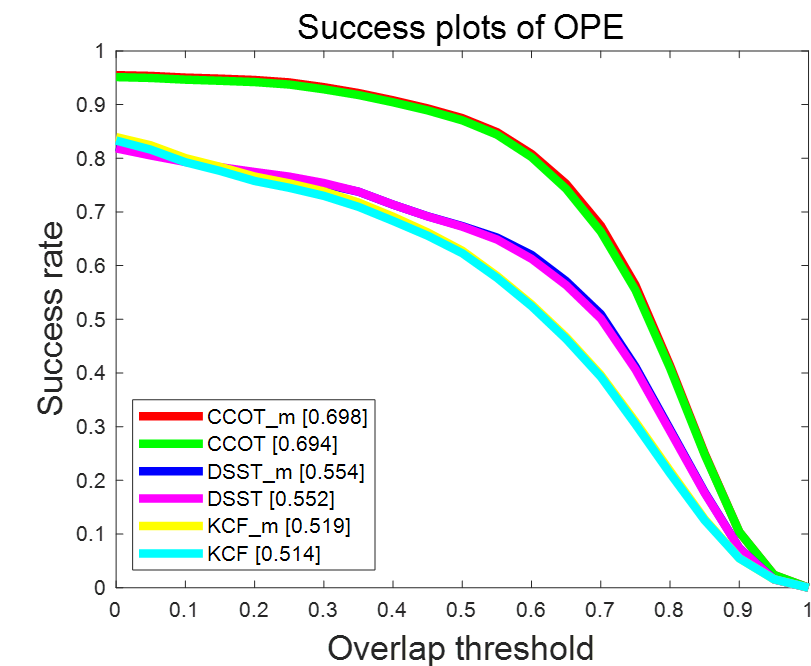}
		\label{Fig.4b}
	}%
	\centering
	\caption{The evaluation of the trackers on the TB50. Where KCF-m, DSST-m and CCOT-m represent the trackers based on KCF, DSST and CCOT, but assisted by the proposed moving target detection algorithm, respectively.}
	\label{Fig.4}
\end{figure}

The evaluation experiments are performed on three state-of-the-art trackers: KCF \citep{kcf}, DSST \citep{dsst} and CCOT\citep{ccot}. KCF is a typical correlation filter tracker. This tracker generates a large number of samples obtained by cyclically shift the region of interest to train the classifier. Then, the similarity between candidate samples and target is calculated by the kernel function, and the sample with the largest similarity will be selected as the new tracking target. Meanwhile, the discrete Fourier transform is used in this tracker to reduce the computational complexity in training and detection. DSST adds a module of scale estimation on the KCF which is realized by scaling the target to different scales for correlation matching. It can still achieve good tracking accuracy when the target scale changes. CCOT is also a state-of-the-art correlation filter tracker. It introduces an interpolation model to transfer the feature maps into continuous spatial domain, which can effectively integrate the multi-resolution feature maps. The application of deep and shallow layers of CNN features makes the tracker have strong classification ability and location resolution. Besides, the continuous function is used to obtain the predicted score of the target, and accurate sub-grid positioning is achieved. These trackers are evaluated on the benchmark TB50\citep{otb2013}, results as shown in Fig.\ref {Fig.4} .

\begin{figure}[htbp]
	\centering
	\subfigure{
		\centering
		\includegraphics[scale=0.35]{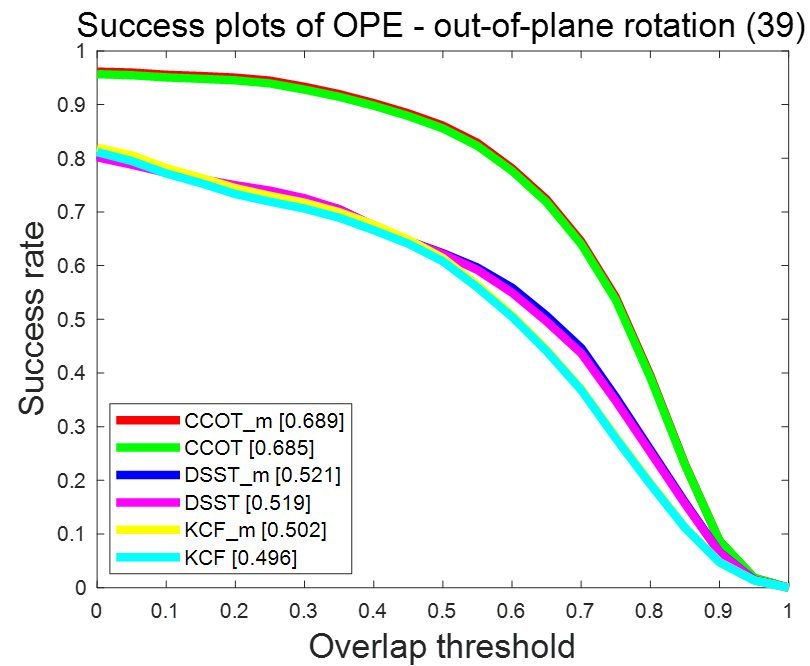}
		\label{Fig.5a}
	}%
	\subfigure{
		\centering
		\includegraphics[scale=0.35]{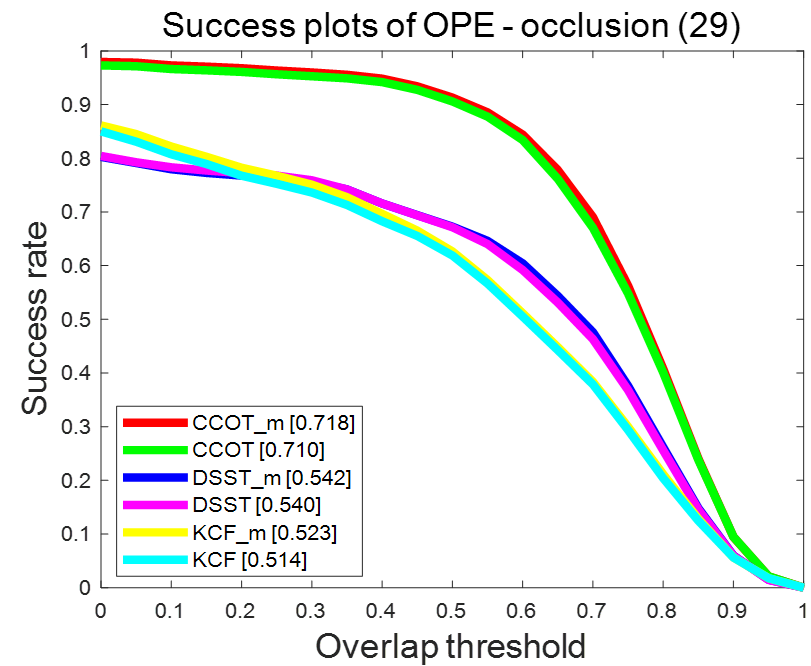}
		\label{Fig.5b}
	}%
	
	\subfigure{
		\centering
		\includegraphics[scale=0.35]{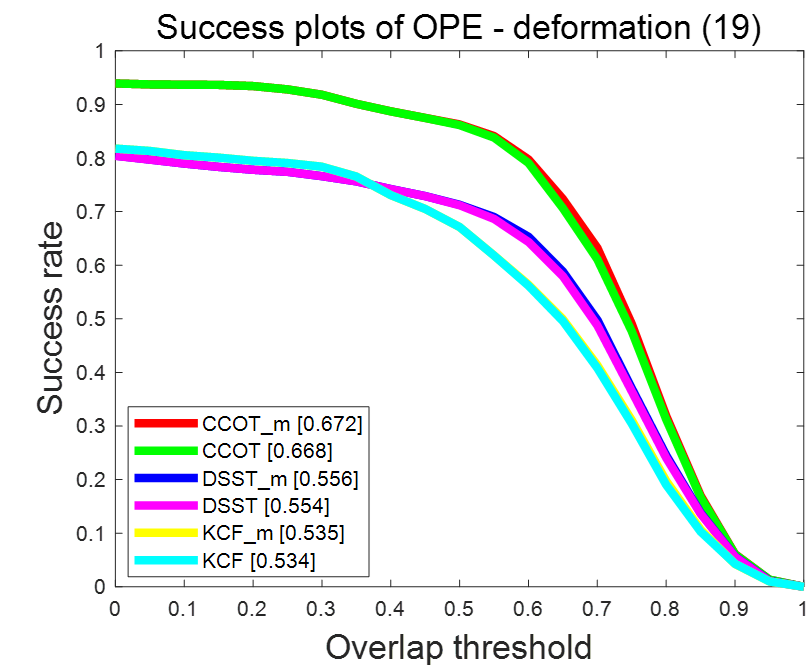}
		\label{Fig.5c}
	}%
	\subfigure{
		\centering
		\includegraphics[scale=0.35]{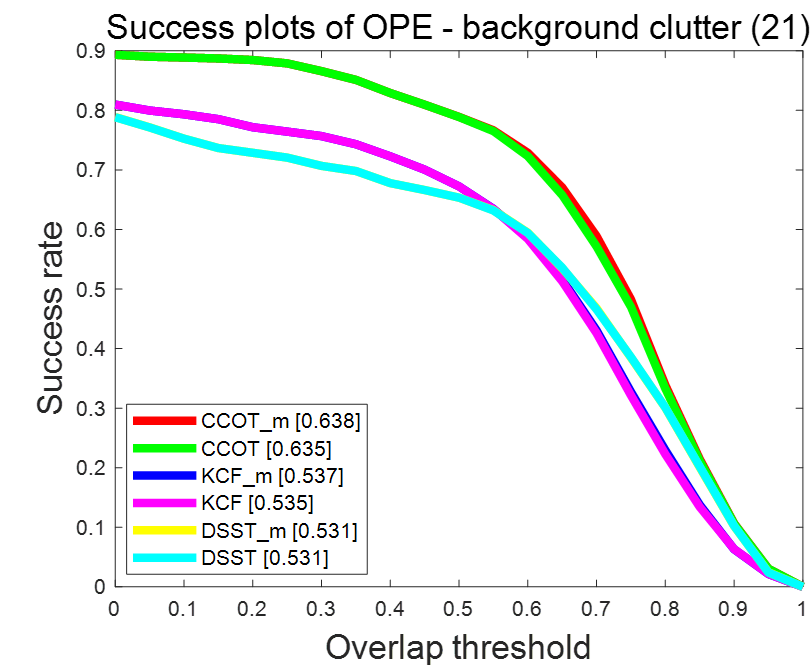}
		\label{Fig.5d}
	}%
	
	\subfigure{
		\centering
		\includegraphics[scale=0.35]{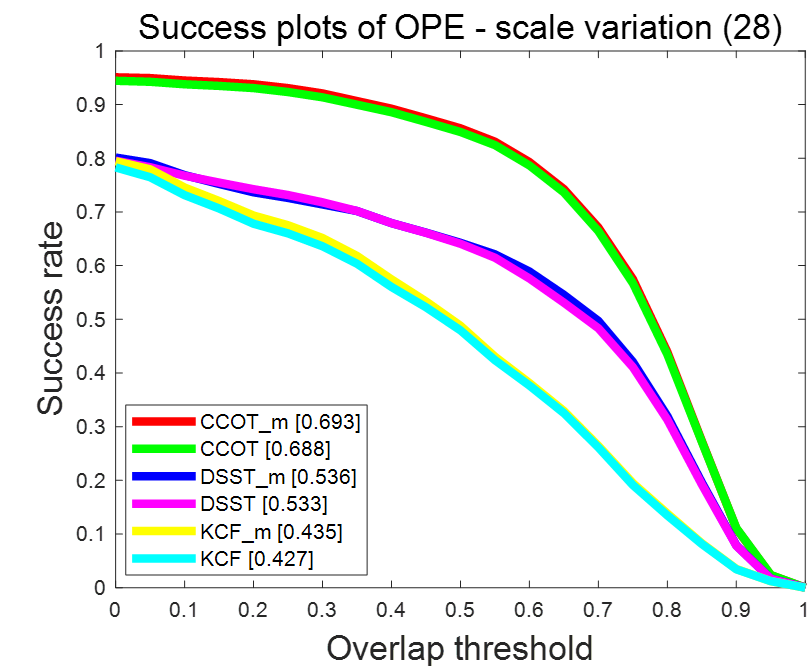}
		\label{Fig.5e}
	}%
	\subfigure{
		\centering
		\includegraphics[scale=0.35]{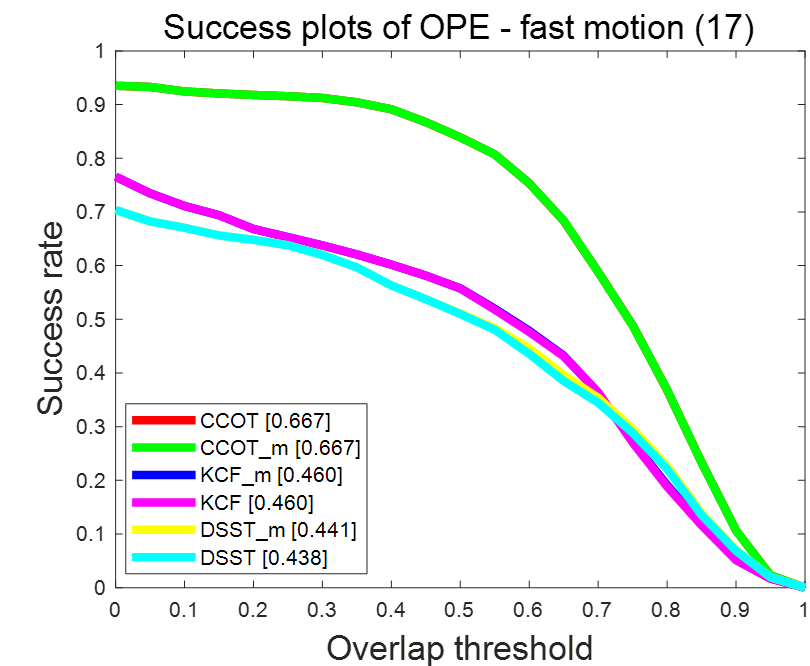}
		\label{Fig.5f}
	}%
	
	\subfigure{
		\centering
		\includegraphics[scale=0.35]{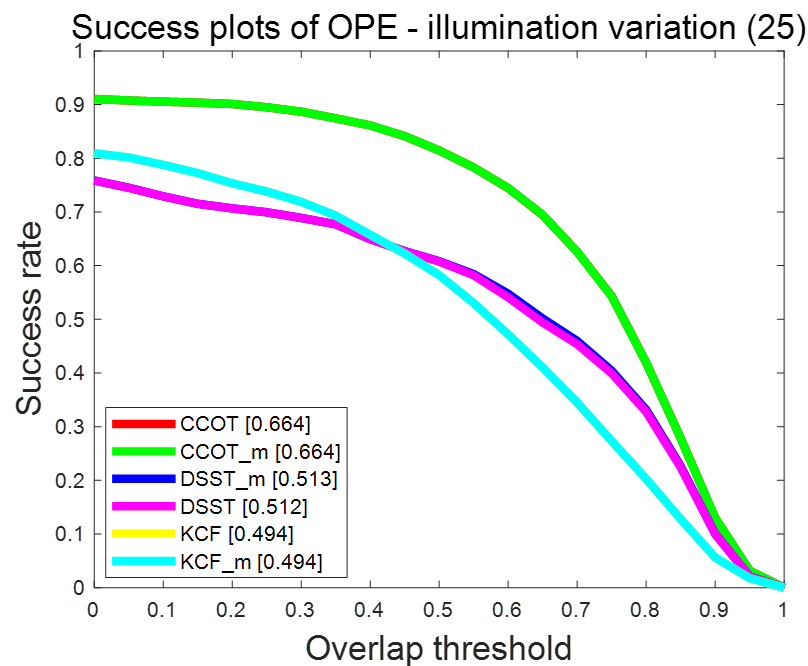}
		\label{Fig.5g}
	}%
	\subfigure{
		\centering
		\includegraphics[scale=0.35]{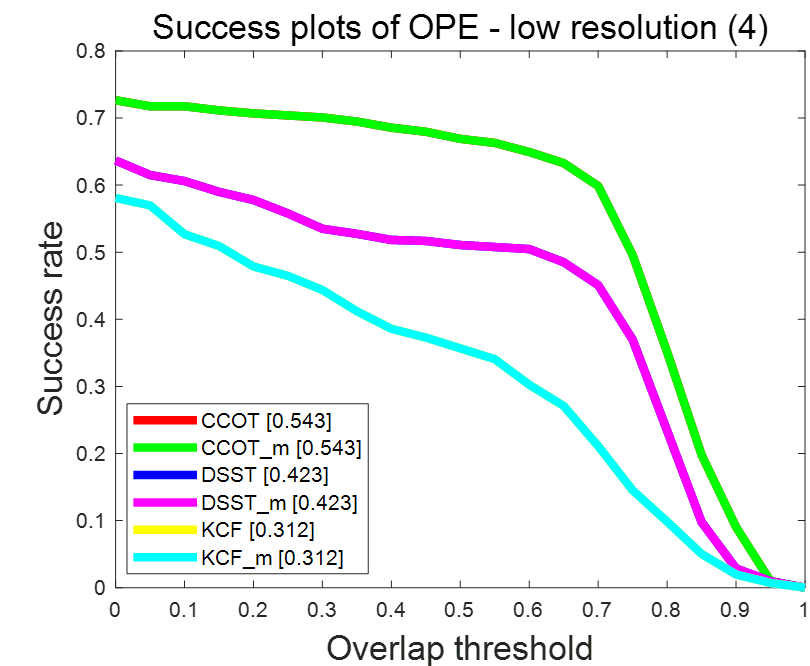}
		\label{Fig.5h}
	}%
	\centering
	\caption{The evaluation of CCOT, DSST, KCF, CCOT-m and DSST-m and KCF-m on challenge scenes of ‘out-of-plane rotation’, ‘occlusion’, ‘deformation’, ‘background clutter’, ‘scale variation’, ‘fast motion’, ‘illumination change’ and ‘low resolution’.}
	\label{Fig.5}
\end{figure}

Different from Fig.\ref {Fig.2}, the lower right corner of precision map in Fig.\ref {Fig.4a} shows the precision at the threshold of 20 pixels. It can be seen from Fig.\ref {Fig.4} that the precision and success rate of the trackers assisted by the proposed moving target detection module have been improved. The precision at the location error threshold of 20 pixels is improved by 0.33\% for CCOT, 0.45\% for DSST and 0.61\% for KCF. The AUC is improved by 0.48\% for CCOT, 0.26\% for DSST and 0.47\% for KCF. Besides, we also evaluate the success rate of these trackers in various challenging scenes, results as shown in Fig.\ref {Fig.5} . It can be found that the success rate of the three trackers have been improved in almost all challenging scenes, with the assistance of the proposed moving target detection module. Especially in occlusion, deformation, out-of-plane rotation and background clutter, where the original trackers perform poorly. The reason is that the texture features of targets are severely damaged in these scenes, which affects the matching accuracy of trackers. The color and motion features make up for the shortcomings of texture features and maintain good classification ability in these scenes. Furthermore, the moving target detection module uses the results already obtained in the tracking process for calculation. Which hardly increases the computational burden and does not affect the real-time performance of the tracking algorithms. Therefore, the proposed moving target detection module can be well applied in trackers to improve tracking performance.

\section{Conclusions}

In this work, we proved the motion feature is important for moving target detection. However, the existing algorithms pay more attention on the semantic characteristics while ignoring the motion characteristics of the moving targets. Therefore, we proposed a class agnostic moving target detection algorithm by adopting the motion feature effectively. The proposed algorithm extracts moving area through deep cnn feature difference. Then, the color probability map and location probability map will be calculated based on the moving area. Next, the target probability can be obtained through multiplying these two maps. And the optimal bounding box of the moving target can be obtained by SGD. Besides, as the existing datasets are not suitable for model-free moving target detection, we proposed a method for generating dataset based on the tracking datasets. We evaluated the proposed algorithm with state-of-the-art algorithms based on the produced dataset. Results show that the proposed algorithm can detect the moving target accurately, which is 15\% more accurate than the state-of-the-art algorithms, and runs in real-time. Furthermore, we also proved that the proposed moving target detection algorithm can be used as a module to assist other computer vision tasks. Its application in tracking algorithms improved the tracking accuracy effectively. However, there are still some shortcomings in this algorithm. It cannot realize multi-targets detection, and difficult to solve the scene with large transformation. Also, it is difficult to achieve end-to-end integration with existing networks. These issues will be addressed in our following researches.

\bibliographystyle{cag-num-names}

\bibliography{mybibfile}

\end{document}